\documentclass[letterpaper]{article} 
\usepackage{aaai2026}  
\usepackage{times}  
\usepackage{helvet}  
\usepackage{courier}  
\usepackage[hyphens]{url}  
\usepackage{graphicx} 
\usepackage{natbib}  
\usepackage{caption} 
\usepackage{algorithm}
\usepackage{algorithmic}

\usepackage{newfloat}
\usepackage{listings}
\DeclareCaptionStyle{ruled}{labelfont=normalfont,labelsep=colon,strut=off} 
\floatstyle{ruled}
\newfloat{listing}{tb}{lst}{}
\floatname{listing}{Listing}
\nocopyright 
\title{Help or Hurdle? Rethinking Model Context Protocol-Augmented Large Language Models}
\author{
    Wei Song\equalcontrib\textsuperscript{\rm 1}\textsuperscript{\rm 2},
    Haonan Zhong\equalcontrib\textsuperscript{\rm 1},
    Ziqi Ding\textsuperscript{\rm 1},
    Jingling Xue\textsuperscript{\rm 1},
    Yuekang Li\textsuperscript{\rm 1}
}
\affiliations{
    \textsuperscript{\rm 1} University of New South Wales, Australia\\
    \textsuperscript{\rm 2} CSIRO's Data61, Australia\\

}

\usepackage{bibentry}

\usepackage{enumitem}
\usepackage{amsmath}
\usepackage{amssymb}
\usepackage{booktabs}

\usepackage{multirow}

\usepackage{tcolorbox}
\usepackage{xcolor}

\newtcolorbox{finding}[1]{
    colback=gray!5!white,
    colframe=gray!40!black,
    coltitle=black,
    colbacktitle=gray!35!white,
    fonttitle=\bfseries,
    title=Finding #1,
    boxrule=0.5pt,
    left=3mm,
    right=3mm,
    top=2mm,
    bottom=2mm
}

\newcommand{\MCP}{\text{MCP}}
\newcommand{\TIR}{\text{TIA}}
\newcommand{\IFA}{\text{IFA}}
\newcommand{\Acc}{\text{Acc}}
\newcommand{\MCPEval}{\textsc{MCPGauge}}

\begin{document}

\pagestyle{plain}

\maketitle

\begin{abstract}
The Model Context Protocol (MCP) enables large language models (LLMs) to access external resources on demand. While commonly assumed to enhance performance, how LLMs actually leverage this capability remains poorly understood.
We introduce \MCPEval{}, the first comprehensive evaluation framework for probing LLM–MCP interactions along four key dimensions: \textit{proactivity} (self-initiated tool use), \textit{compliance} (adherence to tool-use instructions), \textit{effectiveness} (task performance post-integration), and \textit{overhead} (computational cost incurred).
\MCPEval{} comprises a 160-prompt suite and 25 datasets spanning knowledge comprehension, general reasoning, and code generation.
Our large-scale evaluation—spanning six commercial LLMs, 30 \MCP{} tool suites, and both one- and two-turn interaction settings—comprises around 20{,}000 API calls and over USD~6{,}000 in computational cost. 
This comprehensive study reveals four key findings that challenge prevailing assumptions about the effectiveness of \MCP{} integration.
These insights highlight critical limitations in current AI–tool integration and position \MCPEval{} as a principled benchmark for advancing controllable, tool-augmented LLMs.
\end{abstract}
\section{Introduction}
\label{sec:introduction}

The prospect of autonomous AI agents that can seamlessly access diverse tools and data sources has gained considerable traction.
However, this landscape remains fragmented. 
Each tool requires bespoke interface definitions, authentication handling, and execution logic, and function-calling APIs differ across platforms~\cite{MCP-HouXinyi-2025, MCP-Multi-Agent-2025, MCP-Evaluation-Luo-2025}. 
As a result, AI agents are often constrained by static, hard-wired workflows rather than dynamically discovering and orchestrating tools at runtime.

To address these issues, Anthropic released the Model Context Protocol (\MCP) in late 2024~\cite{MCP}. 
\MCP{} aims to streamline AI development and enhance flexibility in managing intricate workflows by standardizing interfaces and enabling agents to dynamically discover, select, and coordinate external services without hard-coded mappings. 
Since its release, \MCP{} has transformed from a protocol into a fundamental building block of AI-driven platforms, 
supported by a vibrant community ecosystem of \MCP{} servers 
that provide connectivity to web search engines, structured databases, file systems, and custom computational APIs.
Instead of requiring LLMs to internalize every piece of knowledge or functionality within their parameters, \MCP{} separates retrieval and execution from generation: 
the LLM issues a ``tool call'' (e.g., a web search or database query), receives back structured snippets (text, tables, code fragments, or numeric data), and then continues reasoning with that injected context. 
Through real-time integration of specialized knowledge from external resources during inference, 
\MCP{} seeks to transform how LLMs generate responses, with the goal of improving accuracy and strengthening reasoning capabilities.

\noindent
\textbf{Research Gap.}
While \MCP{} provides promising infrastructure for tool integration, a significant gap persists between its theoretical benefits and practical usefulness.
The reason is that the final performance on various tasks depends on not only the extra context provided by MCP but also LLMs' capacity to recognize when external tools are needed, execute \MCP{} calls appropriately, and effectively utilize the retrieved information.
Although recent studies have examined \MCP{}'s architecture~\cite{MCP-HouXinyi-2025, MCP-Survey-Singh-2025, MCP-Survey-Ray-2025}, security concerns~\cite{MCP-Attack, MCP-Safety-Review}, and MCP tools' efficiency of requesting resources~\cite{MCP-Evaluation-Luo-2025}, a critical gap remains in understanding how LLMs engage with \MCP{}. 
Existing benchmarks like \MCP{}-RADAR~\cite{MCP-Radar-Gao-2025} evaluate only task performance outcomes without examining the fundamental behavioral aspects of how LLMs recognize tool needs, execute calls, and integrate retrieved information.

\noindent
\textbf{Research Questions.} 
Given this gap, we examine LLM-\MCP{} interaction through four key research questions:

\begin{itemize}[label={}, leftmargin=0pt]

\item 
\textit{RQ1: Do LLMs proactively invoke external tools provided by \MCP{} when such actions could improve task performance, without explicit user instructions?}

\item 
\textit{RQ2: To what extent do LLMs follow explicit user instructions to use \MCP{} tools?}

\item 
\textit{RQ3: How does external context retrieved via \MCP{} tools affect LLM task performance?}

\item
\textit{RQ4: What is the computational overhead, measured in input token increase, associated with \MCP{} tool integration?}

\end{itemize}

\noindent
\textbf{Challenges.}
Answering these research questions presents two key challenges. 
First, the community lacks a clear set of axes for measuring how well models use external tools.
We therefore define four complementary dimensions—\emph{proactivity}, \emph{compliance}, \emph{effectiveness}, and \emph{overhead}, to measure how well a model can proactively invoke a tool (RQ1), how well it obeys explicit directives (RQ2), to what extent the external context helps (RQ3), and how much it costs (RQ4).
Second, existing LLM benchmarks were designed for stand‑alone models and seldom require \MCP{} tool use. 

By addressing the challenges, we introduce \MCPEval, the first systematic evaluation framework for LLM–\MCP{} interaction. 
\MCPEval{} consists of a newly-designed suite of 160 prompts and 25 well-established task datasets.
The prompt suite demands autonomous tool recognition, covering the need for measuring proactivity and compliance.
The task datasets contain knowledge comprehension, general reasoning, and code generation tasks, covering the need for measuring effectiveness and overhead.

We used \MCPEval{} to evaluate six commercial LLMs equipped with 30 \MCP{} tool suites across both one-turn and two-turn dialogue settings.
Our analysis revealed four surprising and insightful findings that challenge common assumptions about the effectiveness of \MCP{} integration:
(1) Most models exhibit minimal proactive use of \MCP{} tools in the first turn, but their behavior improves significantly in two-turn dialogues, suggesting an implicit ``warm-up'' phase is needed before effective tool usage.
(2) Instruction compliance improves only when tool-use directives are embedded within incremental dialogue, indicating a limited ability to follow single-shot commands.
(3) Contrary to expectations, automated \MCP{} access by LLMs reduces accuracy by an average of 9.5\% across the six LLMs on three core task categories, revealing non-trivial friction between retrieved context and the model’s internal reasoning.
(4) \MCP{} integration introduces substantial computational overhead: input-token volume increases by $3.25\times$ to $236.5\times$ across models and tasks.
These findings highlight critical bottlenecks in current LLM–\MCP{} interactions. 
They offer valuable guidance for future research and practical system design, particularly for developing more efficient and controllable \MCP{}-augmented LLM agents.

\noindent
\textbf{Contributions.} We make three major contributions:

\begin{itemize}[leftmargin=*]
\item 
We propose \MCPEval{}, the first comprehensive framework for empirically evaluating LLM interactions with \MCP{} tools. \MCPEval{} includes a 160-prompt suite and 25 datasets covering knowledge comprehension, general reasoning, and code generation tasks.

\item 
We introduce four evaluation dimensions—\textit{proactivity}, \textit{compliance}, \textit{effectiveness}, and \textit{overhead}—to systematically assess an \MCP{}-enabled LLM’s ability to initiate tool use, follow explicit instructions, utilize retrieved context to improve performance, and manage the computational cost of tool integration.

\item 
Our large-scale evaluation—incurring around 20{,}000 LLM API calls and over USD~6{,}000 cost—yields several key insights into current LLM–\MCP{} integration. We find that proactive tool use typically emerges only after a brief conversational context; multi-turn interactions significantly enhance instruction compliance; misalignment between retrieved context and task demands can impair effectiveness; and tool use often incurs substantial input-token overhead.

\end{itemize}

To facilitate open science and future research, we provide the code and raw experiment data as supplementary materials and will open-source them upon paper acceptance.
\section{Background and Related Work}
\label{sec:background}

\begin{figure}[t]
\centering
\includegraphics[width=1\linewidth]{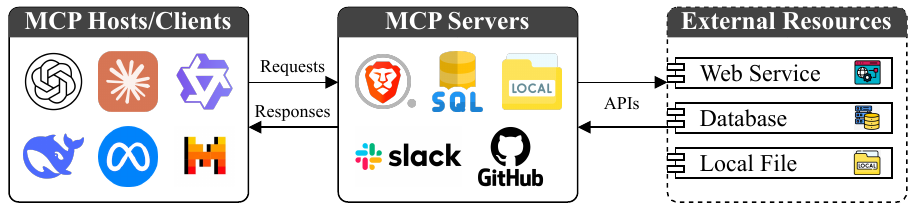}
\caption{Model Context Protocol (\MCP).}
\label{fig:mcp}
\end{figure}

\subsection{Model Context Protocol}
\label{subsec:background_model_context_protocol}
Model Context Protocol (\MCP{}) serves as a communication standard that bridges AI models and external resources. 
The protocol is built on a tripartite architecture featuring \MCP{} hosts, clients, and servers as fundamental building blocks as illustrated in Figure~\ref{fig:mcp}. 
MCP Host: The AI application environment (e.g., Claude Desktop, Cursor IDE) that runs the MCP client and provides the interface for AI-based tasks. 
MCP Client: Acts as the communication bridge between the host and external servers, managing requests, processing responses, and coordinating tool invocations. 
MCP Server: Provides access to external capabilities through three core functionalities: (1) Tools for executing external operations and API calls, (2) Resources for accessing structured and unstructured data sources, and (3) Prompts for reusable workflow templates that optimize AI responses.
Through this structure, \MCP{} ensures controlled and reliable data exchange between AI systems and external services. 
As depicted in Figure~\ref{fig:mcp}, 
\MCP{} operates through a coordinated workflow in which the \MCP{} client receives user prompts, 
collaborates with the \MCP{} server to identify and access relevant external tools, processes the retrieved information and presents the final results to the user. 

\subsection{Related Work}
\label{subsec:related_work}

Multiple surveys \cite{MCP-HouXinyi-2025, MCP-Survey-Singh-2025, MCP-Survey-Ray-2025} have examined the \MCP{} and its emerging ecosystem. 
\citet{MCP-HouXinyi-2025} provide a broad landscape analysis that examines \MCP{}'s architecture and applications while highlighting the absence of systematic evaluation frameworks. 
\citet{MCP-Survey-Singh-2025} focus on standardization efforts for LLM enhancement, 
while \citet{MCP-Survey-Ray-2025} examine \MCP{}'s current applications, challenges, and adoption patterns.

As \MCP{} adoption has expanded, security implications have emerged as a critical concern. 
\citet{MCP-Attack} conduct a comprehensive safety audit revealing major security exploits possible when LLMs interact through \MCP{}, demonstrating that even well-aligned models can be compromised in \MCP{} environments.
This work underscores the importance of systematic evaluation frameworks like ours for understanding LLM behavior in tool-augmented settings. 
Building on these security concerns, \citet{MCP-Safety-Review} propose enterprise-grade security frameworks and mitigation strategies for \MCP{} implementations, addressing the challenges of deploying \MCP{} in production environments while maintaining security standards. 
Complementing this work, \citet{MCP-Attack-Vector} provide a detailed analysis of specific attack vectors and vulnerability exploitation methods in \MCP{} systems.

In parallel with these security investigations, the evaluation of LLMs with the \MCP{} paradigm has taken several distinct directions. 
\citet{MCP-Evaluation-Luo-2025} introduce MCPBench, a systematic evaluation framework for assessing \MCP{} server performance, but does not comprehensively evaluate LLM capabilities in \MCP{} utilization. 
\citet{MCP-Radar-Gao-2025} introduce \MCP{}-RADAR, a benchmark solely focused on measuring performance outcomes through evaluating LLM tool-use capabilities across answer accuracy, tool selection efficiency, computational resource efficiency, parameter construction accuracy, and execution speed.

Although prior research has investigated \MCP{}'s architecture, security, and server performance, they only focus on the \MCP{} side.
As a result, a systematic analysis of interactions between LLMs and \MCP{} is still lacking.
This work addresses that gap by proposing four complementary dimensions for evaluating LLM–\MCP{} interactions and applying them to assess widely used LLMs and \MCP{} tools.
The results reveal key bottlenecks in current \MCP{}-augmented LLM systems and offer valuable insights for future research.

\begin{figure*}[t]
\centering
\includegraphics[width=1\linewidth]{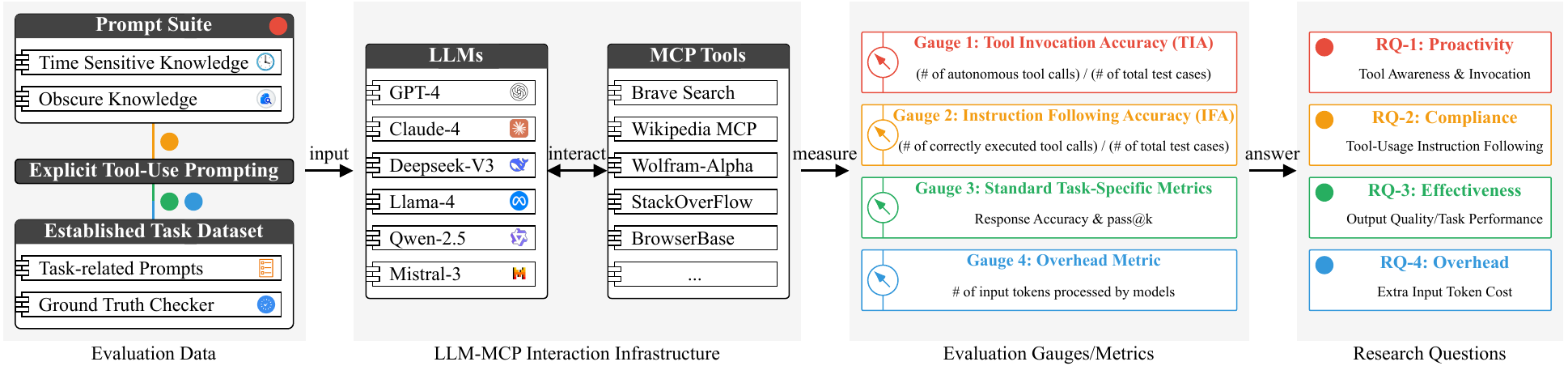}
\caption{Framework of \MCPEval{}.}
\label{fig:evaluation_framework}
\end{figure*}

\section{\MCPEval{} Design}
\label{sec:mcp_benchmark}

\MCPEval{} systematically assesses LLM-MCP interactions across four critical dimensions
through a tailored 160-prompt suite and 25 datasets covering knowledge comprehension, general reasoning, and code generation tasks, across four \MCP-specific dimensions: \emph{proactivity (RQ1)}, \emph{compliance (RQ2)}, \emph{effectiveness (RQ3)}, and \emph{overhead (RQ4)}.
Figure~\ref{fig:evaluation_framework} shows an overview of \MCPEval{}.

\subsection{Evaluation Dimensions}
\label{subsec:evaluation_dimensions}

We introduce four \MCP-specific evaluation dimensions, \textit{proactivity}, \textit{compliance}, \textit{effectiveness}, and \textit{overhead} to comprehensively examine LLM-MCP interactions:

\begin{itemize}[leftmargin=*]
\item 
\textit{Proactivity} is quantified as the proportion of task scenarios 
where 
LLMs correctly identify the need for external tools and successfully invoke them. It 
measures whether LLMs demonstrate self-directed awareness of their knowledge limitations, for example, needing real-time web search for current events, 
and proactively initiate appropriate MCP tool calls without explicit instruction (\textbf{RQ1}).

\item 
\textit{Compliance} measures LLMs' adherence to explicit user instructions regarding MCP tool usage. When users instruct LLMs to use MCP tools by mentioning \MCP{} generally (e.g., ``use \MCP{} to get today's weather'') or by specifying particular tools (e.g., ``use web search tool to find current stock prices'')—we evaluate whether LLMs follow instructions to execute the requested tool calls (\textbf{RQ2}).

\item 
\textit{Effectiveness} quantifies the improvement in output quality when LLMs utilize MCP-retrieved context. We assess how external context impacts performance by comparing outputs with and without \MCP{} integration (\textbf{RQ3}).

\item 
\textit{Overhead} captures the computational cost introduced by \MCP{} integration. Concretely, we compare the total number of input tokens that LLMs must process with and without \MCP{}, thereby quantifying the additional resource footprint that \MCP‑augmented reasoning imposes (\textbf{RQ4}).
\end{itemize}

\subsection{Design Rationale}
\label{subsec:evaluation_approaches}

\subsubsection{Dimension‑specific Design}
For \emph{proactivity}, we construct scenarios where \MCP{} calls are necessary for solving the problems but not explicitly mentioned by user commands. 
In order to ensure the need of external tool invocations for task completion, we design to craft queries requiring  time-sensitive knowledge (e.g., current weather, recent news, and today's stock price) or obscure knowledge (e.g., information about some small companies).
The rationale of the design is to make sure that the required knowledge is not used for training the LLMs.
Therefore, external context provided by \MCP{} calls are needed.

To evaluate \emph{compliance},
we should include in the prompts explicit instructions for \MCP{} tool usage such as ``use \MCP{} calls to find/search/retrieve ...''. 

For \emph{effectiveness}, we adopt a controlled comparison approach, evaluating LLM performance on identical tasks with and without \MCP{} tools.
Therefore, the ground-truth for performance evaluation is important.
Task datasets with correct/standard answers as ground-truth are needed for evaluation.

To quantify \emph{overhead}, we propose to tally the full input‑token workload, including the user prompt, system instructions, and any retrieved context that each LLM must process during inference.

\subsubsection{Conversation-depth Settings} 
We evaluate proactivity and compliance under two dialogue depths, \emph{one-turn} and \emph{two-turn} to gauge how conversational context influences tool use. The one-turn setting records the model’s very first response, mirroring the common single‑shot scenario in which users expect an immediate answer. 
The two‑turn setting appends a follow‑up query that mimics real‑world dialogues, allowing users to refine their request or supply additional instructions.
Based on our preliminary experiments, we found that the performance improvement of three or more turns is marginal.
Therefore, we focus on one and two-turn conversations and leave the exploration of more conversation turns as future work.

\subsection{Evaluation Datasets}
\label{subsec:benchmark_dataset}
To evaluate LLM–\MCP{} interactions along the four dimensions, we construct a 160‑prompt suite for assessing proactivity and compliance, and employ 25 benchmark datasets to measure effectiveness and overhead.

For proactivity, existing LLM benchmarks~\cite{EvalPlus, AGI-Eval} are inadequate because they typically include task specifications that do not require external tools. 
We therefore construct a 160-prompt suite that deliberately withholds any tool mention yet requires \MCP{} calls for task completion according to the design rationale discussed previously.
For compliance, we reuse the same 160 prompts but insert explicit instructions for invoking \MCP{} calls into the prompts.
Details about the prompts we have crafted are in the supplementary materials.

For effectiveness and overhead, we draw on 25 existing well-established LLM benchmarks.
The reason is that they provide ground-truth for performance evaluation.
The benchmark datasets cover three task domains --- knowledge comprehension, general reasoning, and code generation, as detailed below:

\begin{itemize}[leftmargin=*]
\item
\textit{Knowledge Comprehension (KC)} We utilize two datasets from HellaSwag \cite{HellaSwag}: ActivityNet, which provides video captions, and WikiHow, which consists of how-to articles for completion tasks.

\item 
\textit{General Reasoning (GR)} We incorporate 21 datasets from AGI-Eval \cite{AGI-Eval}, including 
test datasets (SAT Math, SAT English, LSAT Analytical Reasoning, LSAT Logical Reasoning, LSAT Reading Comprehension), 
mathematical reasoning tasks (AQUA-RAT, Math), 
logical reasoning datasets (LogiQA-EN, LogiQA-ZH), 
comprehensive academic assessments from Gaokao covering various subjects (Chinese, English, Geography, History, Biology, Chemistry, Physics, Math QA, Math Cloze),
and legal reasoning tasks (JEC-QA-KD, JEC-QA-CA).

\item 
\textit{Code Generation (CG)} We include two datasets from EvalPlus \cite{EvalPlus}: HumanEval and MBPP, which focus on Python programming tasks such as algorithm implementation, code debugging, and test case generation.

\end{itemize}

It is worth noting that all 25 datasets have their own user prompts and queries.
Therefore, the 160-prompt suite is not used here.
Instead, similar to how we modify the prompts for evaluating compliance, we insert explicit instructions for calling \MCP{} tools into the prompts in these benchmarks to increase the frequency of LLM-\MCP{} interactions during the task solving processes.
Moreover, since LLMs may not choose to use \MCP{} tools even when explicitly instructed, we only collect and analyze the tasking solving instances where \MCP{} calls are made by LLMs for the purpose of studying the impact of \MCP{} calls on performance and overhead.

\subsection{Evaluation Metrics}
\label{subsec:evaluation_metrics}

We define formal metrics to quantitatively evaluate LLM–MCP interactions along the four proposed dimensions:

\begin{itemize}
    \item \textbf{Tool Invocation Accuracy (TIA)}: Measures the proportion of test cases where the LLM autonomously invokes an appropriate MCP tool without being explicitly instructed. Formally,
    \[
    \text{TIA} = \frac{N_{\text{autonomous}}}{N_{\text{total}}}
    \]
    where $N_{\text{autonomous}}$ is the number of test cases in which the LLM proactively initiates a correct MCP tool call, and $N_{\text{total}}$ is the total number of tool-dependent test cases.

    \item \textbf{Instruction Following Accuracy (IFA)}: Measures the LLM’s adherence to explicit tool-use instructions embedded in prompts. Formally,
    \[
    \text{IFA} = \frac{N_{\text{compliant}}}{N_{\text{total}}}
    \]
    where $N_{\text{compliant}}$ denotes the number of test cases where the LLM correctly executes the instructed MCP tool call, and $N_{\text{total}}$ is the total number of instruction-containing prompts.

    \item \textbf{Effectiveness (Acc, pass@k)}: Assesses the improvement in task performance when MCP-retrieved context is integrated. We use domain-appropriate metrics:
    \begin{itemize}
        \item For knowledge and reasoning tasks:
        \[
        \text{Accuracy} = \frac{N_{\text{correct}}}{N_{\text{total}}}
        \]
        where $N_{\text{correct}}$ is the number of task responses matching ground-truth answers.
        \item For code generation tasks, we adopt the unbiased pass@k metric as defined in \cite{PassK}.
    \end{itemize}

    \item \textbf{Overhead (Token Cost)}: Quantifies the computational cost introduced by MCP integration in terms of input token volume. Formally,
    \[
    \text{Overhead Ratio} = \frac{T_{\text{with-MCP}}}{T_{\text{without-MCP}}}
    \]
    where $T_{\text{with-MCP}}$ and $T_{\text{without-MCP}}$ represent the total input tokens (including user prompt, instructions, and retrieved context) with and without MCP, respectively.
\end{itemize}

\section{Experiments}
\label{sec:experiments}

\subsection{Experiment Setup}
\label{subsec:experiment_setup}

\begin{itemize}[leftmargin=*]

\item 
\textit{Large Language Models} We evaluate six leading commercial LLMs: 
GPT-4, Claude-4, DeepSeek-V3, Llama-4, Qwen-2.5, and Mistral-3. 
This selection encompasses diverse model architectures, 
providing comprehensive coverage of current LLM-MCP interactions.

\item 
\textit{MCP Tools} We integrate 30 official \MCP{} tool suites, providing diverse information retrieval domains, including general web search \MCP{} tools (e.g., \texttt{brave\_web\_search}), 
general knowledge-focused retrieval \MCP{} services (e.g., \texttt{wikipedia\_mcp}), 
mathematical problems-related searching \MCP{} tools (e.g., \texttt{wolfram-alpha}), 
and programming questions-related retrieval \MCP{} tools (e.g., \texttt{stackoverflow\_mcp}).
We manually verify that the 30 \MCP{} tools are enough for covering all the need for the tasks in the 160-prompt suite and the 25 benchmark datasets.
A list of the selected tools are provided in the supplementary material.

\item
\textit{Computation Platform} All experiments are conducted on one NVIDIA RTX 3090 GPU with 24GB RAM.

\end{itemize}

\subsection{Results}
\label{subsec:results}

\begin{figure}[t]
\centering
\includegraphics[width=1\linewidth]{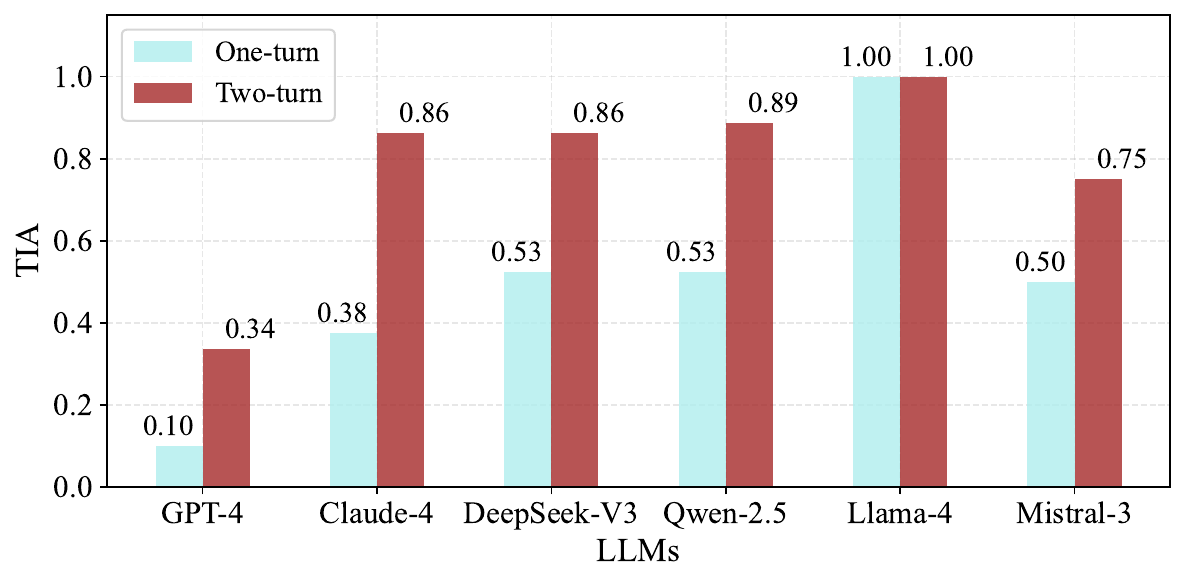}
\caption{\textbf{Proactivity (RQ1).} Autonomous invocation of \MCP{} tools by six LLMs on tool‑dependent prompts, compared between one‑turn and two‑turn dialogue settings.}
\label{fig:evaluation_RQ1}
\end{figure}

\begin{table*}[t]
\centering
\setlength{\tabcolsep}{1.6pt}
\begin{tabular}{llccccccccccc}
\toprule
\multirow{2}{*}{Model} & \multirow{2}{*}{Setting} & \multicolumn{2}{c}{KC(\Acc$\uparrow$)} & \multicolumn{7}{c}{GR(\Acc$\uparrow$)} & \multicolumn{2}{c}{CG(pass@$k$$\uparrow$)} \\
\cmidrule(lr){3-4} \cmidrule(lr){5-11} \cmidrule(lr){12-13}
 &   & ActivityNet & WikiHow & SAT & LSAT & Gaokao & LogiQA & JEC & AQUA & MATH & HumanEval & MBPP \\
\midrule
\multirow{2}{*}{GPT-4} 
 & Without MCP &   0.80  &  0.94 &  0.90  &  0.87 &  0.77 &  0.77 &  0.60 & 0.88 &  0.39 & 0.92 & 0.78  \\
 & With MCP    &   0.79  &  0.94 &  0.89  &  0.79 &  0.79 &  0.74 &  0.65 & 0.88 &  0.65 & 0.91 & 0.80      \\
\midrule
\multirow{2}{*}{Claude-4} 
 & Without MCP &   0.86  &  0.99 &  0.87  &  0.87 & 0.85      &  0.87     &  0.70   &  0.90   & 0.70  &  0.92  & 0.79  \\
 & With MCP    &   0.82  &  0.97 &  0.92  &  0.82 & 0.79      &  0.78     &  0.69   &  0.85   & 0.60  &  0.86 &  0.67     \\
\midrule
\multirow{2}{*}{DeepSeek-V3} 
 & Without MCP &  0.77   & 0.97   & 0.91  &  0.85 & 0.85      &  0.81     &  0.80     & 0.84   & 0.58   &  0.88     &   0.74    \\
 & With MCP    &   0.78  & 0.92   & 0.89  &  0.80 & 0.78      &  0.76     &  0.65     & 0.69   & 0.59   &  0.86     &   0.72     \\
\midrule
\multirow{2}{*}{Qwen-2.5} 
 & Without MCP &  0.76   &  0.97  & 0.85  &  0.73 & 0.80      &  0.71     & 0.70     & 0.75    &  0.60     &    0.82   &   0.79    \\
 & With MCP    &  0.70   &  0.92  & 0.79  &  0.66 & 0.62      &  0.72     & 0.60     & 0.73    &  0.40     &    0.61   &  0.38      \\
\midrule
\multirow{2}{*}{Llama-4} 
 & Without MCP & 0.72  &  0.91    & 0.81 &  0.73  &  0.64     &  0.66     & 0.48    &  0.82    &  0.72     &    0.79   &   0.75    \\
 & With MCP    & 0.73  &  0.89    & 0.80 &  0.65  &  0.59     &  0.60     & 0.48    &  0.71    &  0.55     &    0.59   &   0.62   \\
\midrule
\multirow{2}{*}{Mistral-3} 
 & Without MCP &  0.74  & 0.93    & 0.88  &  0.78  &  0.81     &  0.70     & 0.57    & 0.87   & 0.58  &   0.86    &   0.64    \\
 & With MCP    &  0.81  & 0.91    & 0.75  &  0.73  &  0.65     &  0.72     & 0.46    & 0.88   & 0.59  &   0.81    &   0.53    \\
\bottomrule
\end{tabular}
\caption{\textbf{Effectiveness (RQ3).} Effectiveness evaluation of the six commercial LLMs with and without MCP integration across three critical task domains: Knowledge Comprehension (KC), General Reasoning (GR), and Code Generation (CG).}
\label{table:RQ3_Effectiveness}
\end{table*}


\noindent
\textbf{Proactivity.}  We evaluate six leading commercial LLMs' autonomous \MCP{} tool recognition through a tailored 160-prompt suite designed to necessitate real-time information access, e.g., current weather, recent news, without explicit \MCP{} tool usage instructions. 
We measure Tool Invocation Accuracy (\TIR)—the proportion of cases where models successfully recognize tool necessity and initiate appropriate \MCP{} calls across one-turn and two-turn dialogue settings.

As shown in Figure \ref{fig:evaluation_RQ1}, most LLMs exhibit limited autonomous tool recognition in one-turn settings, but conversational engagement yields remarkable improvements in tool invocation capabilities: 
GPT-4 achieves a 240.0\% enhancement, Claude-4 demonstrates 128.9\% improvement, DeepSeek-V3 shows 62.3\% growth, Qwen-2.5 exhibits 67.9\% increase, and Mistral-3 displays 50.0\% enhancement—with Llama-4 maintaining consistent perfect performance across both settings. 
In detail, one-turn scenarios reveal substantial variation in proactive capabilities: GPT-4 demonstrates minimal tool awareness with \TIR{} of 0.10, Qwen-2.5 shows moderate autonomous recognition at 0.53, Claude-4 exhibits limited proactivity at 0.38, while DeepSeek-V3 and Mistral-3 achieve comparable moderate proactivity at 0.53 and 0.50, respectively, and Llama-4 reaches optimal \TIR{} of 1.0. 
Conversational interactions markedly enhance active \MCP{} tool usage, with Claude-4, DeepSeek-V3, and Qwen-2.5 achieving tool invocation accuracy of 0.86, 0.86, and 0.89, respectively. 
GPT-4 progresses to 0.34, Mistral-3 advances to 0.75, while Llama-4 sustains perfect proactivity at \TIR{} of 1.0. Therefore, adding a second conversational turn wakes up the models’ ability to notice when a tool is needed, turning a mostly unused capacity into active, proactive behaviour for nearly all LLMs.

\begin{figure}[t]
\centering
\includegraphics[width=1\linewidth]{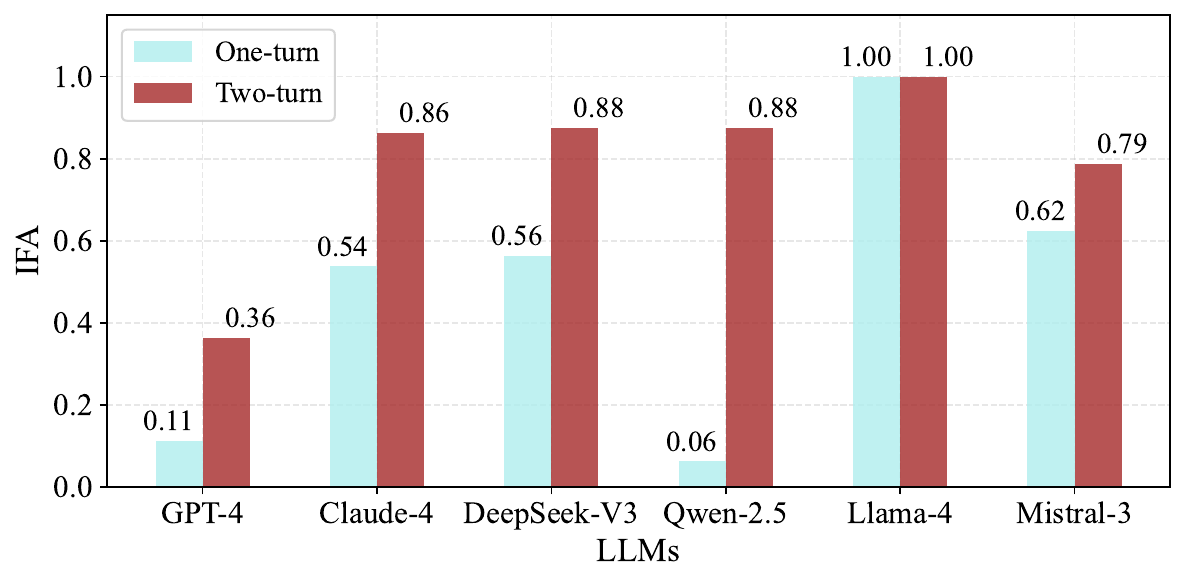}
\caption{\textbf{Compliance (RQ2).} Adherence of six LLMs to explicit MCP tool‑use instructions in prompts, compared between one‑turn and two‑turn dialogue settings.}
\label{fig:evaluation_RQ2}
\end{figure}

\begin{finding}{1}
Most LLMs require cognitive ``warm-up'' to recognize \MCP{} tool necessity—they lack proactive \MCP{} tool awareness in one-turn scenarios while achieving substantially improved proactivity in two-turn interactions.
\end{finding}




\noindent
\textbf{Compliance.} We next assess these six leading commercial LLMs' instruction-following capabilities for explicit \MCP{} tool usage through a tailored 160-prompt suite that contains direct tool usage directives. We measure Instruction Following Accuracy (\IFA)—the proportion of cases where models successfully follow explicit tool usage instructions—across one-turn and two-turn conversation scenarios.

As illustrated in Figure \ref{fig:evaluation_RQ2}, one-turn conversation scenarios elicit weak compliance from every model except Llama-4, yet a single follow-up turn dramatically improves instruction following accuracy by 227.3\% for GPT‑4, 59.3\% for Claude‑4, 57.1\% for DeepSeek‑V3, an exceptional 1366.7\% for Qwen‑2.5, and 27.4\% for Mistral‑3—while Llama‑4 remains perfect instruction following accuracy of 1.00 in both settings. Specifically, 
in one-turn conversation scenarios, GPT-4 shows poor compliance with \IFA{} of 0.11, while Qwen-2.5 demonstrates notably poor instruction-following at 0.06 \IFA. Claude-4 and DeepSeek-V3 achieve moderate \IFA{} of 0.54 and 0.56, respectively. Mistral-3 and Llama-4 demonstrate higher \IFA{} performance at 0.62 and 1.0, respectively. In contrast, two-turn conversations reveal substantially improved compliance across most models, with Claude-4, DeepSeek-V3, and Qwen-2.5 all reaching similar high \IFA{} levels of 0.86, 0.88, and 0.88, respectively. Mistral-3 shows compliance improvement to 0.79, while Llama maintains perfect compliance at 1.0. 
Thus, brief conversational scaffolding transforms explicit MCP directives from largely ignored hints into reliably executed actions, with significant improvements for nearly all models.

\begin{finding}{2}
Most LLMs lack genuine instruction-following capability—apparent \MCP{} tool usage compliance emerges through conversational context-building rather than immediate directive execution.
\end{finding}



\noindent
\textbf{Effectiveness.} Beyond proactivity and compliance, we also evaluate \MCP's impact on LLM's effectiveness across three core task domains: knowledge comprehension, general reasoning, and code generation using well-established LLM benchmarks. 
We conduct comparative experiments between MCP-augmented and standalone LLM configurations, measuring performance using standard accuracy (\Acc) \cite{AGI-Eval} for knowledge comprehension and reasoning tasks, and pass@$k$ \cite{PassK} for code generation tasks to assess the code's functional correctness.

As shown in Table \ref{table:RQ3_Effectiveness}, instead of enhancing model performance, integration of \MCP{} generally degrades LLM effectiveness
across the three task domains, averaging 9.5\% performance decline across the six LLMs and the three task domains, when \MCP{} tools are employed compared to standalone operation. This suggests that external information may introduce noise or conflicting signals that interfere with models' internal reasoning processes, rather than providing beneficial supplementary context. In detail, on knowledge comprehension  (KC) tasks, most LLMs exhibit slight accuracy degradation, averaging 1.4\% decline across all LLM-dataset pairs with \MCP{} integration. 
Across ActivityNet and WikiHow datasets, the accuracies of GPT-4, Llama-4, Claude-4, DeepSeek-V3, and Qwen-2.5 degrade by 0.6\%, 0.4\%, 3.3\%, 1.9\%, and 6.5\%, respectively, while Mistral-3 improves by 3.7\%. These results indicate that external information introduces conflicting contextual cues that interfere with models' parametric knowledge in these tasks. 
On reasoning tasks, LLMs exhibit substantial accuracy decline averaging 10.2\% across all LLM-dataset combinations. 
Across SAT, LSAT, Gaokao, LogiQA, JEC, AQUA, and MATH, while GPT-4 improves by 4.2\%, 
Mistral-3, Llama-4, Claude-4, DeepSeek-V3, and Qwen-2.5 degrade by 11.4\%, 15.2\%, 7.1\%, 12.8\%, and 18.6\%, respectively. 
These patterns indicate that external information systematically disrupts complex reasoning processes across most model architectures.  
Code generation tasks reveal the most severe performance degradation, averaging 17.0\% decline in pass@$k$ when provided with external context through \MCP{} calls. 
Across HumanEval and MBPP, although GPT-4 improves by 1.4\%, Mistral-3, Llama-4, Claude-4, DeepSeek-V3, and Qwen-2.5 degrade by 22.2\%, 18.4\%, 9.9\%, 5.1\%, and 48.1\%, respectively. This pronounced vulnerability suggests that external information particularly interferes with the precise cognitive processes required for code synthesis.

\begin{finding}{3}
Contrary to expectations, automated \MCP{} integration results in performance degradation across three major task domains, rather than yielding improvements. This finding suggests that current LLMs are not yet capable of effectively leveraging external information retrieved via \MCP{} tools in an autonomous manner.
\end{finding}

\noindent
\textbf{Overhead.} 
To quantify the computational cost of \MCP{} integration, we measure the overhead incurred by each LLM. Specifically, for each model and task domain, we record the total number of input tokens processed with and without \MCP{} calls. This comparison captures the increase in prompt length, and by extension, the additional computational burden introduced by context retrieved via \MCP{}.


\begin{finding}{4}
MCP integration imposes a substantial computational overhead—input‐token volume grows by $3.25\times$ to $236.5\times$ across the six LLMs and three task domains.
\end{finding}

As demonstrated in Table \ref{table:RQ4_Token_Cost}, across the six models and three task domains, integrating \MCP{} context expands the input token volume by $3.25\times$ to $236.5\times$, showing that \MCP{} tool integration carries a steep computational cost. Specifically, on knowledge comprehension (KC) tasks, the input token load increases by $3.25\times$, $236.5\times$, $24.5\times$, $53.3\times$, $53.8\times$, and $23.0\times$ for GPT-4, Claude-4, DeepSeek-V3, Qwen-2.5, Llama-4, and Mistral-3, respectively, on general reasoning (GR) tasks, the cost surges by $24.1\times$, $81.4\times$, $38.4\times$, $22.5\times$, $57.4\times$, and $22.6\times$, and on code generation (CG) tasks, the token consumption escalates by $135.7\times$, $212.7\times$, $117.5\times$, $53.9\times$, $96.1\times$, and $110.8\times$.

\begin{table}[t]
\centering
\begin{tabular}{llccc}
\toprule
Model & Setting & KC & GR & CG \\

\midrule
\multirow{2}{*}{GPT-4} 
 & Without MCP &  0.04   &  0.60 &  0.06  \\
 & With MCP    &  0.17  & 14.45  &  8.20  \\
\midrule
\multirow{2}{*}{Claude-4} 
 & Without MCP &  0.04   & 0.70  & 0.07   \\
 & With MCP    &  9.46   & 57.00  & 14.89    \\
\midrule
\multirow{2}{*}{DeepSeek-V3} 
 & Without MCP &  0.04   & 0.53   & 0.06   \\
 & With MCP    &  0.98  &  20.34  & 7.05   \\
\midrule
\multirow{2}{*}{Qwen-2.5} 
 & Without MCP & 0.04    &  0.58  & 0.07   \\
 & With MCP    &  2.13   & 13.07   & 3.77   \\
\midrule
\multirow{2}{*}{Llama-4} 
 & Without MCP & 0.05  & 0.66     & 0.08  \\
 & With MCP    & 2.69  & 37.86     & 7.69   \\
\midrule
\multirow{2}{*}{Mistral-3} 
 & Without MCP & 0.04   & 0.62    & 0.06  \\
 & With MCP    & 0.92   & 13.99    & 6.65  \\
\bottomrule
\end{tabular}
\caption{\textbf{Overhead (RQ4).} Total input tokens (in millions) processed by the six LLMs with and without MCP tool calls across three task domains, knowledge comprehension (KC), general reasoning (GR), and code generation (CG).}
\label{table:RQ4_Token_Cost}
\end{table}

\section{Discussion}
\label{sec:discussion}

Our results overturn the assumption that LLMs naturally work well with \MCP{} tools: proactivity lags, explicit directives are often ignored, and added context can hurt accuracy and significantly increase the costs. These pinpoint clear gaps in interface design, instruction following, and context merging—that future AI‑tool integration must address.

\noindent
\textbf{Cognitive ``Warm‑up'' and Its Deployment Implications.} Our experiments show that nearly every model invokes tools far more often after a short follow‑up turn than on the initial query. 
This ``warm‑up'' effect suggests that current LLMs do not instantly recognize gaps in their parametric knowledge; they need extra conversational context before switching from internal reasoning to external tool use. 
For time‑sensitive applications, e.g., fetching live stock prices or weather—such hesitation can translate into stale or incomplete answers, because users typically expect useful tool calls in the very first response. 
Two practical lessons follow. First, interface designers should not assume one‑shot queries will trigger tool usage; adding a lightweight system follow‑up (``Feel free to consult available tools'') can markedly boost productivity. 
Second, LLM developers may need architectural change, such as explicit uncertainty detectors or dedicated tool‑planner heads to reduce dependence on multi‑turn scaffolding and enable immediate, context‑aware tool invocation.

\noindent
\textbf{Instruction-Following Limitations.}
When explicit tool directives are placed in the very first prompt, models comply only infrequently—far below the compliance levels they reach after a second conversational turn. Even when the need for a tool is made explicit, the instruction is often ignored. This indicates that current LLMs parse imperative phrasing less reliably than incremental conversational cues. For deployment, rigid one‑shot directives alone are insufficient; coupling them with lightweight follow‑ups may be necessary to ensure immediate tool invocation.

\noindent
\textbf{External‑Information Integration Bottleneck.}
Incorporating \MCP{}-retrieved context leads to performance degradation rather than improvement, with an average drop of 9.5\%. These results indicate that current LLMs struggle to autonomously benefit from \MCP{} calls.
Bridging this gap will likely require more effective filtering mechanisms or model-level adaptations to ensure that only truly relevant information contributes to the final output.
Consequently, incorporating additional programmatic logic—such as agentic system design—may enhance LLMs' ability to interact effectively with \MCP{} tools.

\noindent
\textbf{Computational‑Cost Trade‑off.}
Tool calls come with a steep price: integrating MCP context expands the input token by $3.25\times$ to $236.5\times$, inflating FLOPs, memory footprint, latency, and API usage fees. Such expansion can quickly exhaust context windows, slow first‑token latency, and raise serving costs by orders of magnitude—particularly for code‑generation workloads where the longest snippets are already near the model’s limit. In practice, system designers may need token‑budget guards, relevance pruning, or client‑side caching to ensure that the benefit of an external lookup outweighs the extra compute and billing overhead.

\section{Conclusion}
We present \MCPEval{}, the first end-to-end framework for evaluating how LLMs interact with the \MCP{}. Leveraging a 160-prompt suite and 25 established datasets, \MCPEval{} assesses four key dimensions: proactivity, compliance, effectiveness, and overhead.
Our evaluation of six commercial LLMs reveals that proactive tool use rarely occurs on the first turn, explicit one-turn directives are often ignored, and retrieved context frequently degrades accuracy while inflating input-token budgets by up to $236.5\times$.
These findings highlight fundamental limitations in current LLM-\MCP{} integration and establish \MCPEval{} as a robust benchmark for developing more reliable and cost-efficient tool-augmented LLMs.

\bibliography{aaai2026}

\end{document}